# Solving Games with Functional Regret Estimation

**Kevin Waugh**[†]
waugh@cs.cmu.edu

**Dustin Morrill**[*]
morrill@ualberta.ca

**J. Andrew Bagnell**[†]
dbagnell@ri.cmu.edu

**Michael Bowling**[*]
mbowling@ualberta.ca

**School of Computer Science**[†]
Carnegie Mellon University
5000 Forbes Ave
Pittsburgh, PA 15213 USA

**Department of Computing Science**[*]
2-21 Athabasca Hall
University of Alberta
Edmonton, AB T6G 2E8 Canada

## Abstract

We propose a novel online learning method for minimizing regret in large extensive-form games. The approach learns a function approximator online to estimate the regret for choosing a particular action. A no-regret algorithm uses these estimates in place of the true regrets to define a sequence of policies.

We prove the approach sound by providing a bound relating the quality of the function approximation and regret of the algorithm. A corollary being that the method is guaranteed to converge to a Nash equilibrium in self-play so long as the regrets are ultimately realizable by the function approximator. Our technique can be understood as a principled generalization of existing work on abstraction in large games; in our work, both the abstraction as well as the equilibrium are learned during self-play. We demonstrate empirically the method achieves higher quality strategies than state-of-the-art abstraction techniques given the same resources.

## Introduction

Online learning in sequential decision-making scenarios has wide ranging applications (*e.g.*, online path planning (Awerbuch and Kleinberg 2004), opponent exploitation (Southey, Hoehn, and Holte 2009), and portfolio optimization (Hazan et al. 2006)). While many such applications have considerable structure, often this structure is lost when formulating the application into a traditional problem specification such as a multi-armed bandit or extensive-form game. Since regret-minimizing algorithms usually require computational resources and learning time that grows with the problem representation size, losing this structure often makes the resulting problem intractable. Currently, the remedy is for the practitioner to abstract the original domain based on structural features into a problem with a tractable representation size (*e.g.*, using domain knowledge to preselect a small number of likely-orthogonal choices in a multi-armed bandit).

Consider the popular testbed of poker. The human-played game is highly-structured with a compact set of rules. Its unstructured extensive-form game representation, however, with nodes in the game tree for every possible sequence of states is anything but compact (*viz.*, two-player, no-limit Texas hold'em has $8.2 \cdot 10^{160}$ such sequences (Johanson 2013, p. 12)). Even specialized equilibrium-finding algorithms have no hope of solving this game in its extensive-form representation. Here, much of the recent progress has employed abstraction, grouping together similar situations and actions based on domain knowledge (Gilpin, Sandholm, and Sorensen 2007; Johanson 2007).

Using abstraction as a preliminary step to capture all of the possible structure of the domain is unsatisfying. It can place limitations on how structural features are exploited, *e.g.*, in extensive-form games the resulting abstraction must be discrete and result in perfect recall. Furthermore, it requires the practitioner to have foresight as to what the learner might face when it is actually making decisions.

In this paper, we present a new algorithm for online learning in adversarial sequential decision-making scenarios that makes use of structural features of the domain *during learning*. In particular, we employ an online regressor from domain features to approximate regrets incurred by the regret minimizing online algorithm. This simple change allows the algorithm to employ more general forms of abstraction, and can change and tune the abstraction to both the game and its solution. We prove a bound on our approach relating the accuracy of the regressor to the regret of the algorithm and demonstrate its efficacy in a simplified poker game.

## Regression Regret-Matching

Let us begin by presenting a new, yet simple, algorithm for the standard online learning framework. Let $A$ be the set of $N$ **actions** or **experts**. On each time step $t \in [T] \equiv \{1, \ldots, T\}$, the online algorithm chooses $x^t \in \Delta_A$, a distribution over the experts, then observes $u^t$, a bounded reward vector where $\|u^t\|_\infty \leq L^2$. The algorithm receives reward $(x^t \cdot u^t)$ and then updates its prediction.

An algorithm's **external regret** compares its performance to the best expert in hindsight. To be **no-regret** is to have regret grow sublinearly in $T$,

$$R^{\text{ext}} = \max_{x^* \in \Delta_A} \sum_{t=1}^{T} (x^* \cdot u^t) - (x^t \cdot u^t) \in o(T).$$

That is, its average regret goes to zero in the worst-case.

**Algorithm 1** Regret-matching with Regret Regression

$X \leftarrow [], y \leftarrow []$
**for** $t \in [T]$ **do**
    $f \leftarrow \text{TrainRegressor}(X, y)$
    **for** $a \in A$ **do**
        $z(a) \leftarrow f(\varphi(a))$
    **end for**
    $x^t \propto \max\{0, z^t\}$
    Observe $u^t$
    **for** $a \in A$ **do**
        $X \leftarrow X \cup \varphi(a)$
        $y \leftarrow y \cup (u^t(a) - (x^t \cdot u^t))$
    **end for**
**end for**

A simple algorithm with this property is regret-matching.

**Definition 1** (Regret-matching). *Define the vector of **cumulative regret** as*

$$R^T \equiv \sum_{t=1}^{T} u^t - (x^t \cdot u^t)e$$

*where $e$ is the vector of all ones. Choose $x^t \propto \max\{0, R^{t-1}\}$.*

**Theorem 1** (from Hart & Mas-Colell (2000)). *Regret-matching is no-regret. In particular, $R^{ext} \leq \sqrt{LNT}$.*

We often have structure between the available actions, *e.g.*, when an action corresponds to betting a particular amount or choosing an acceleration. It is common to model such situations by discretizing the range of available actions and using a single action per interval. In this case, the structure linking actions is completely lost after this abstraction.

In more complicated stateful settings, we can describe each action $a \in A$ with a **feature vector**, $\varphi(a) \in \mathcal{X}$. In poker, there are numerous ways to quantify the strength of a particular hand. For example, its expected value against different possible opponent holdings as well as potential-aware statistics, like the variance of the expected value. Current techniques use eight to ten such statistics to describe the portion of the state space representing the player's hand.

To reduce the action space in these settings, prior work uses unsupervised clustering techniques on the actions in feature space. This can essentially be thought of as an informed discretization of the space. The hope is that actions with similar features can be collapsed, or made indistinguishable, without incurring too much loss (Shi and Littman 2002). We use this feature vector to retain the structure *during learning*. To reduce to the standard framework, we can simply use an indicator feature for each action.

Algorithm 1 contains the pseudocode for our approach. It employs an arbitrary mean-squared-error minimizing regressor to estimate the cumulative regret. In particular, the training procedure is expected to produce $f$ where $f(\varphi(a)) \approx R^t(a)/t$, or equivalently, the algorithm's regret estimate is $\tilde{R}^t(a) = tf(\varphi(a))$.

We consider the error of our regressor with the $l_2$ distance between the approximate and true cumulative regret vectors: $\|R^t - \tilde{R}^t\|_2$. Note that this quantity is related to the representational power and recall of the regressor. In particular, if we cannot represent the true cumulative regret then it will be non-zero due to this bias. The regret of Algorithm 1 can then be bounded in terms of this representational power.

**Theorem 2.** *If for each $t \in [T]$, $\|R^t - \tilde{R}^t\|_2 \leq \epsilon$, then regression regret-matching has regret bounded by $\sqrt{TNL} + 2T\sqrt{L}\epsilon$.*

The proof is in the appendix in the supplemental material.

This is a worst-case bound. That is, so long as $\epsilon$ is small regression regret-matching cannot be much worse than normal regret-matching. It is possible that there are cases where regression regret-matching can do better than regret-matching if the structure is favorable.

The first subtlety we must note is that the regressor minimizes the mean-squared error of the immediate regrets, *i.e.*

$$\{(\varphi(a), u^i(a) - u^i \cdot x^i) \mid \forall a \in A, i \in [t]\}$$

forms the training set. If the hypothesis class can represent the regrets, then $tf(\varphi(a)) = R^t(a)$ obtains the minimum mean-squared error. Note that this error is likely not zero even in the realizable case!

In words, the theorem states that if the error of the regressor decreases like $O(1/T)$ then the algorithm obtains a $O(\sqrt{T})$ regret bound. Note that the cumulative regret can grow like $O(T)$, so a fixed $\epsilon$ across time implies a decreasing error rate. The algorithm remains no-regret so long as the bias goes to zero and recall goes to one, *i.e.*, it is asymptotically unbiased. If the bias is constant then there is a constant term in the regret bound. When solving a game, this constant term constitutes the error introduced by a lossy abstraction.

Note that it is sufficient to make the estimator asymptotically unbiased by including indicator features for each action with proper regularization. In a sense, this allows the algorithm to move from an estimate of the true regrets as time increases, *i.e.*, as the true regrets stabilize.

Next, we aim to use our algorithm for sequential decision-making and to relate it to current abstraction techniques. Before we can accomplish this, we must review the extensive-form game framework.

## Extensive-form Games

A **two-player zero-sum extensive-form game** is a tuple $\Gamma = (\mathcal{H}, p, \sigma_c, \mathcal{I}, u)$ (Osborne and Rubinstein 1994). The set of **histories**, $\mathcal{H}$, form a tree rooted at $\phi \in \mathcal{H}$, the empty history. $A(h)$ is the set of actions available at $h \in \mathcal{H}$ and $ha \in \mathcal{H}$ is a child of $h$. The subset $\mathcal{Z} \subseteq \mathcal{H}$ is the set of **terminal histories**, *i.e.*, if $z \in \mathcal{Z}$ then $A(z)$ is empty. $p : \mathcal{H} \setminus \mathcal{Z} \to \{1, 2, c\}$ is the **player choice function** that determines if nature or a player is to act at any given history. For all histories $h \in \mathcal{H}$ where chance acts (that is, all $h$ where $p(h) = c$), $\sigma_c(\cdot|h) \in \Delta_{A(h)}$ is a probability distribution over available chance actions that defines **nature's strategy**. The **information partition**, $\mathcal{I} = \mathcal{I}_1 \cup \mathcal{I}_2$, separates histories into **information sets** such that all histories in an information set are indistinguishable to the acting player and have the same set of actions. Finally, $u : \mathcal{Z} \to \mathbb{R}$ is the game's **utility function**. At a terminal history, $z \in \mathcal{Z}$, $u_1(z) = u(z)$ is the reward to the first player and $u_2(z) = -u(z)$ the reward to the second. That is, the gains of one player are the loses to the other and the rewards sum to zero.

A **behavioral strategy for player $i$**, $\sigma_i \in \Sigma_i$, defines a probability distribution at all information sets where player $i$ acts. That is, if $I \in \mathcal{I}_i$, then $\sigma_i(\cdot|I) \in \Delta_{A(I)}$. We call a tuple of strategies $(\sigma_1, \sigma_2)$ a **strategy profile**. Let $\pi^\sigma(z)$ be the probability of reaching $z$ by traversing the tree using $\sigma$ from the root. Let $\pi^\sigma_{-i}(z)$ be the probability of reaching $z$ using $\sigma$ assuming player $i$ takes actions to reach $z$ with probability one. Finally, let $\pi^\sigma(h, z)$ be the probability of reaching $z$ using $\sigma$ while starting at history $h$ instead of the root ($\pi^\sigma(h,z)$ is zero if $h$ is not an ancestor of $z$). With these definitions we can write the **expected utility to player $i$** under profile $\sigma$ as $u_i(\sigma) = \sum_{z \in \mathcal{Z}} \pi^\sigma(z) u_i(z)$.

A strategy profile is an **$\varepsilon$-Nash equilibrium** if
$$u_1(\sigma_1, \sigma_2) + \varepsilon \geq u_1(\sigma'_1, \sigma_2), \text{ and} \quad \forall \sigma'_1 \in \Sigma_1$$
$$u_2(\sigma_1, \sigma_2) + \varepsilon \geq u_2(\sigma_1, \sigma'_2). \quad \forall \sigma'_2 \in \Sigma_2$$

That is, if neither player can benefit by more than $\varepsilon$ by deviating from $\sigma$ unilaterally. In a two-player zero-sum game, a strategy belonging to a Nash equilibrium is minimax optimal so it attains the highest utility against an optimal opponent. Thus, it is safe to play an equilibrium strategy.

## Counterfactual Regret Minimization

Counterfactual regret minimization (CFR) is an algorithm for computing an $\varepsilon$-equilibrium in extensive-form games (Zinkevich et al. 2008). It employs multiple no-regret online learners, customarily instances of regret-matching, minimizing counterfactual regret at every information set.

The **counterfactual utility** for taking action $a \in A(I)$ from information set $I \in \mathcal{I}_i$ at time $t$ is defined as
$$u_i^t(a|I) = \sum_{h \in I} \sum_{z \in \mathcal{Z}} \pi^t_{-i}(h) \pi^t(ha, z) u_i(z).$$

In words, it is the expected utility for player $i$ taking action $a$ at information set $I$ assuming they play to reach $I$, take action $a$, and then follow their current policy thereafter.

Using counterfactual utility, we define **immediate counterfactual regret** and **cumulative counterfactual regret** as
$$r_i^t(a|I) = u_i^t(a|I) - \sum_{a' \in A(I)} \sigma_i^t(a'|I) u_i^t(a'|I), \text{ and}$$
$$R_i^T(a|I) = \sum_{t=1}^T r_i^t(a|I),$$

respectively. Each information set is equipped with its own no-regret learner that updates and maintains its cumulative regret. It is the cumulative regret from time $t$ that defines the player's strategy at time $t+1$.

Typically, regret-matching is the no-regret learner of choice with CFR. It defines the player's policy at time $t$ as
$$\sigma_i^t(a|I) \propto \max\{0, R_i^{t-1}(a|I)\}.$$

The following two theorems show counterfactual regret minimization in self-play converges to an equilibrium.

**Theorem 3.** *If two no-regret algorithms in self-play each have no more than $\varepsilon$ external regret, $R_i^{T,ext} \leq \varepsilon$, then the average of their strategies form a $2\varepsilon/T$-Nash equilibrium.*

**Theorem 4** (from Zinkevich et al. (2008))**.** *If regret-matching minimizes counterfactual regret at each information set for $T$ iterations, then $R_i^{T,ext} \leq |\mathcal{I}_i|\sqrt{LNT}$, where $L^2$ is the maximum possible utility and $N$ is the maximum number of actions at any information set.*

Counterfactual regret minimization is the large-scale equilibrium-finding algorithm of choice. Though other algorithms boast better asymptotic rates (Gilpin et al. 2007), CFR more rapidly converges to an acceptable solution in practice. Furthermore, it is simple, highly parallelizable and amenable to sampling techniques (Lanctot et al. 2009) that dramatically reduce computational requirements.

## Regression CFR

We are now ready to put the pieces together to form our new regret algorithm for sequential decision-making scenarios: Regression CFR (RCFR). The algorithm is simple. We will minimize counterfactual regret at every information set using estimates of the counterfactual regret that comes from a single common regressor shared across all information sets. The common regressor uses features, $\varphi(I, a)$, that are a function of the information-set/action pair. This allows the regressor to generalize over similar actions and similar situations in building its regret estimates.

As with CFR, we can derive a regret bound, which in turn implies a worst-case bound on the quality of an approximate equilibrium resulting from self-play.

**Theorem 5.** *If for each $t \in [T]$ at every information set $I \in \mathcal{I}_i$, $\|R^t(\cdot|I) - \tilde{R}^t(\cdot|I)\|_2 \leq \epsilon$, then RCFR has external regret bounded by $|\mathcal{I}_i|\sqrt{TNL} + 2T\sqrt{L}\epsilon$.*

The proof combines Theorem 2 together with the CFR convergence proof of Zinkevich et al. (2008).

Now that we have presented RCFR, let us examine how it relates to modern abstraction techniques.

### Relationship to Abstraction

As noted in the introduction, often the problem we wish to solve, when viewed without any structure, is intractably large. For example, representing a behavioral strategy profile in no-limit Texas Hold'em[1] requires $8.2 \cdot 10^{160}$ entries (Johanson 2013, p. 12). To surmount this, we typically first distill the game to a tractably-sized abstract game. Then, after solving it, we map its solution into the full game. The hope is that the abstract game and its solution roughly maintain the important strategic aspects of the full game. Though this turns out to be false (Waugh et al. 2008), it appears to not occur with any significance in the large games of interest (Johanson et al. 2011; Bard et al. 2013).

---
[1] We consider the game with 50-100 blinds with 20,000 chip stacks as played in the Annual Computer Poker Competition.

The common way to abstract a game is to simply group together similar information sets (Gilpin, Sandholm, and Sorensen 2007; Johanson 2007). That is, we take situations in the full game that the player can differentiate and make them indistinguishable in the abstract game. If the two situations have similar behavior in equilibrium, then at the very least we have not lost representational power in doing so. This form of abstraction provides a function $f$ mapping full game information sets to abstract game information sets.

To create such an abstraction requires some notion of similarity and compatibility of information sets, and a clustering algorithm, like k-means. Ultimately, the size of the abstract game is a function of the number of clusters—how the information sets collapse together. Let us assume that we have a function $\varphi : \mathcal{I} \times A \to \mathcal{X}$ that maps information-set/action pairs to domain-specific features in space $\mathcal{X}$. We can use such a function to define the similarity of compatible information sets $I$ and $I'$ by, for example, the cumulative inner product, $\sum_{a \in A} \langle \varphi(I, a), \varphi(I', a) \rangle$, which can be passed to $k$-means to cluster the information sets (often subject to constraints such as preserving perfect recall).

In order to compare the approach to RCFR, consider a single iteration of the counterfactual regret update in both the original game and the abstract game where the players' strategies are fixed. In particular, we have

$$\tilde{r}_i^t(a|I^{\text{abstract}}) = \sum_{I^{\text{full}} \in f^{-1}(I^{\text{abstract}})} r_i^t(a|I^{\text{full}}).$$

That is, the regret at information set $I^{\text{abstract}}$ in the abstract game, $\tilde{r}_i^t(\cdot|I^{\text{abstract}})$ is the sum of the regrets in the full game of all information sets that map to it, $f^{-1}(I^{\text{abstract}})$. Taking this view-point in reverse, using CFR on the abstract game is operationally equivalent to solving the full game where we maintain and update the regrets $\tilde{r}_i^t(\cdot|f(I^{\text{full}}))$, i.e., we approximate the true regrets with a tabular regressor $r_i^t(a|I^{\text{full}}) \approx \tilde{r}_i^t(a|f(I^{\text{full}}))$. Thus, abstraction can be thought of as a special-case of RCFR with a particular choice for how to approximate the cumulative counterfactual regret.

RCFR is, of course, not restricted to tabular regressors and so can capture structure that goes beyond traditional abstraction. For example, using linear or kernel regression provides a sort of "soft" abstraction. That is, the regret of two distinct (in feature space) actions can effect one another without making the two completely indistinguishable, which is the only option available to traditional "hard" abstraction.

Unlike traditional abstraction where $f$ is chosen a priori, RCFR is also able to learn the structure of the abstraction online upon seeing actual data. And since the regressor is trained after each time step, RCFR can effectively re-abstract the game on-the-fly as necessary. Furthermore, the abstraction is informed by the game *in a way that is compatible with the learner*. Imperfect information games have an interesting property in that portions of the game that are never reached in optimal play have strategic effects. For example, if one player was to behave suboptimally, the other might benefit by deviating into an otherwise undesirable space. Without the possibility of deviating there may be no way to punish the poor play. Practically, this presents as a rather annoying hurdle for abstraction designers. In particular, seemingly ideal abstractions tailored perfectly to the structure of an equilibrium may actually lead to poor solutions due to missing important, but seemingly unnecessary, parts of the strategy space. RCFR avoids all of this as the regressor tunes the abstraction to the current policy, not the solution.

## Experimental Results

In order to illustrate the practicality of RCFR, we test its performance in Leduc Hold'em, a simplified poker game. Our goal is to compare the strategies found by RCFR with varying regressors to strategies generated with conventional abstraction techniques. In addition, we examine the iterative behaviour of RCFR compared to CFR.

### Leduc Hold'em

Leduc Hold'em is a poker game based on Kuhn poker (Southey et al. 2005). It provides a good testbed as common operations, like best response and equilibrium computations, are tractable and exact.

The game has two betting rounds, the preflop and flop. At the beginning of the game both players ante a single chip into the pot and are dealt a single private card from a shuffled deck of six cards—two jacks, two queens and two kings. Then begins the preflop betting round where the first player can either check or bet. If the first player checks, passing their turn, then the second player can end the betting round by checking as well, or continue by betting. When facing a bet; the player can raise by placing two chips into the pot; call by matching the bet in the pot and ending the round; or fold by forfeiting the pot to the opponent. There is a maximum of two wagers per round, *i.e.*, one bet and one raise. A single public card is dealt face up for both players to see at the beginning flop. If the flop betting ends without either player folding, a showdown occurs and the player with the best hand takes the pot. A player that pairs, *i.e.*, their card matches the public card, always has the best hand no matter the rank of the paired card or the opponent's card. If neither player pairs, the one with the highest rank card wins. In the event of a tie, the pot is split. The size of a wager preflop is two chips, and is doubled to four chips on the flop.

Leduc Hold'em has 672 sequences. At equilibrium, the first player is expected to lose 0.08 chips per hand. We show results in milliblinds/antes per hand (mb/h), a thousandth of a chip, *i.e.*, optimally the first player loses 80 mb/h.

### Features and Implementation

We use a regression tree aiming to minimize mean-squared error as our regressor. When training, we examine all candidate splits on a single feature and choose the one that results in the best immediate error reduction. The data is then partitioned according to this split and we recursively train both sets. If the error improvement at a node is less than a threshold, or no improvement can be made by any split, a leaf is inserted that predicts the average. It is this error threshold that we manipulate to control the complexity of the regressor—the size of the tree. All the training data is kept between iterations, as in Algorithm 1.

Eight features were chosen such that the set of features would be small, thus allowing fast regression tree training, which is done on every RCFR iteration, but still descriptive enough to have a unique feature expansion for every sequence: **1)** the expected hand strength ($E[HS]$), or the probability of winning the hand given the available information, and marginalized over a uniform distribution of opponent hands and possible future board card; **2)** the rank of the board card, or zero on the preflop; **3)** the pot size; **4)** the pot relative size of the wager being faced, or zero if not facing a wager; **5)** the number of actions this hand; **6)** the number of wagers this hand; **7)** an indicator on whether or not the next action would be a fold action; and **8)** the pot relative size of the wager that would be made by taking the next action, or zero if the next action would be a check or call.

Features **(1)** and **(2)** refer to private and public card information, **(3)** through **(6)** are public chip and action information, while **(7)** and **(8)** fully describe the next potential action in the sequence[2].

### Experiments

We evaluate the empirical performance of RCFR here according to three metrics: **1)** convergence properties, **2)** exploitability, and **3)** one-on-one competitions.

Strategies were computing using RCFR and four different error threshold values. Each threshold was chosen so that RCFR's regressor would have similar complexity to that of a conventional abstraction, or that of the full game. In Leduc Hold'em, a typical abstraction is one that groups together cards on the preflop and only distinguishes between pairing and non-pairing board cards on the flop. These hand-crafted abstractions are analogous to the $E[HS]$ based abstractions commonly used in Texas Hold'em. Abstractions are denoted, for example, J.QK to describe the abstraction that can distinguish between a jack and a queen or king, but cannot distinguish between a queen and king. The remaining three abstractions are then JQK, JQ.K, and J.Q.K. One may also note that J.Q.K is a strict refinement of the other three abstractions, and J.QK and JQ.K both are strict refinements of JQK. To generate strategies, chance sampling CFR (Zinkevich et al. 2007) was run for 100000 iterations to solve each abstract game.

Each different RCFR strategy is denoted, for example, RCFR-22%, to describe RCFR using a regressor 22% the size of a strategy in the full game. RCFR-22%, RCFR-47%, and RCFR-66% correspond to JQK, J.QK/JQ.K, and J.Q.K, respectively, in terms of complexity. RCFR-96% corresponds to FULL, which denotes no abstraction, and it was made by setting the error threshold to zero, so the regressor was free to split on every feature and become as large as the full game. RCFR and CFR were run for 100000 iter-ations to generate the set of RCFR strategies and a FULL strategy, respectively.

---

[2] No explicit check/call feature is necessary because it is implicitly encoded by features **(7)** and **(8)** in combination. The action would be a check/call if and only if both are zero (the action would not be a fold nor a wager, and the game has only three action types).

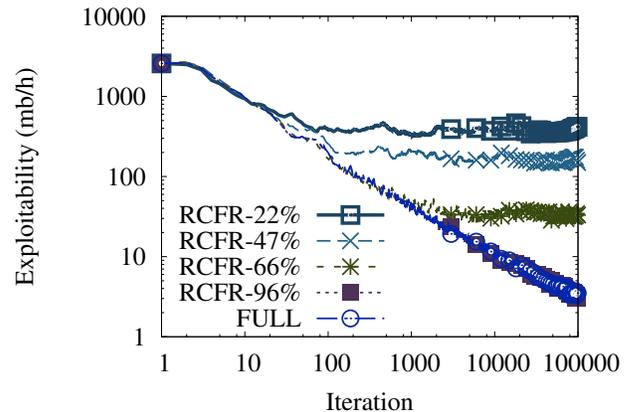

Figure 1: Convergence of RCFR using various error thresholds (complexity limitations) compared with CFR on the unabstracted game (FULL).

**Convergence** Figure 1 shows that all RCFR strategies improve at the same rate as an unabstracted CFR strategy (FULL) until a plateau is reached, the height of which is determined by the error threshold parameter of that RCFR instance. These plateaus are essentially the exploitability cost incurred by estimating regrets instead of computing and storing them explicitly. Larger thresholds, reducing regressor complexity, incur a greater cost and thus have a higher plateau. As expected, when the error threshold is set to zero, as in the case of RCFR-96%, RCFR's progression mimics unabstracted CFR for the full set of 100000 iterations.

**Exploitability** Figure 2 shows that RCFR, given complexity restrictions equivalent to those of conventional abstractions, finds significantly less exploitable strategies. RCFR-66%'s regressor is 2% smaller than the size of the J.Q.K abstract game, yet J.Q.K is sixteen times more exploitable! The closest corresponding strategies in terms of exploitability are RCFR-47% and JQ.K where JQ.K is *only* three and a half times more exploitable.

Another useful practical property of RCFR is that it appears to avoid non-monotonicities that have been observed in hand-crafted abstractions (Waugh et al. 2008). That is, increasing the complexity of the regressor appears to improve the full game exploitability of the resulting strategy.

**One-on-one Competitions** Table 1 is the one-on-one competition crosstable between each of the agents. Against almost every opponent, each RCFR variant outperforms its corresponding strategy. The exceptions, for example, JQ.K wins 100.70 mb/h against RCFR-22% while RCFR-47% wins only 97.61 mb/h against this same opponent, are small margins. In addition, each RCFR strategy defeats or, in the case of RCFR-96%, ties its counterparts. RCFR-22% and RCFR-47% even win against larger abstract strategies J.QK and J.Q.K, respectively. Dividing the agents into an RCFR

|  | RCFR-22% | JQK | RCFR-47% | J.QK | JQ.K | RCFR-66% | J.Q.K | RCFR-96% | FULL | Mean |
|---|---|---|---|---|---|---|---|---|---|---|
| RCFR-22% |  | $319.50 \pm 1.81$ | $-97.61 \pm 1.31$ | $58.25 \pm 1.45$ | $-100.70 \pm 1.26$ | $-123.70 \pm 1.30$ | $-88.90 \pm 1.34$ | $-135.89 \pm 1.34$ | $-139.90 \pm 1.33$ | $-38.62 \pm 1.39$ |
| JQK | $-319.50 \pm 1.81$ |  | $-453.35 \pm 1.85$ | $-405.96 \pm 1.97$ | $-363.66 \pm 1.70$ | $-452.82 \pm 1.85$ | $-369.79 \pm 1.83$ | $-495.46 \pm 1.85$ | $-486.06 \pm 1.86$ | $-418.33 \pm 1.84$ |
| RCFR-47% | $97.61 \pm 1.31$ | $453.35 \pm 1.85$ |  | $140.32 \pm 1.44$ | $14.19 \pm 1.27$ | $-34.91 \pm 1.26$ | $18.07 \pm 1.29$ | $-44.91 \pm 1.26$ | $-47.98 \pm 1.27$ | $74.47 \pm 1.37$ |
| J.QK | $-58.25 \pm 1.45$ | $405.96 \pm 1.97$ | $-140.32 \pm 1.44$ |  | $-107.66 \pm 1.35$ | $-147.86 \pm 1.42$ | $-94.31 \pm 1.45$ | $-149.31 \pm 1.41$ | $-156.26 \pm 1.43$ | $-56.00 \pm 1.49$ |
| JQ.K | $100.70 \pm 1.26$ | $363.66 \pm 1.70$ | $-14.19 \pm 1.27$ | $107.66 \pm 1.35$ |  | $-46.71 \pm 1.26$ | $-33.16 \pm 1.30$ | $-53.90 \pm 1.27$ | $-56.77 \pm 1.27$ | $45.91 \pm 1.33$ |
| RCFR-66% | $123.70 \pm 1.30$ | $452.82 \pm 1.85$ | $34.91 \pm 1.26$ | $147.86 \pm 1.42$ | $46.71 \pm 1.26$ |  | $33.65 \pm 1.26$ | $-10.17 \pm 1.25$ | $-9.41 \pm 1.24$ | $102.51 \pm 1.35$ |
| J.Q.K | $88.90 \pm 1.34$ | $369.79 \pm 1.83$ | $-18.07 \pm 1.29$ | $94.31 \pm 1.45$ | $33.16 \pm 1.30$ | $-33.65 \pm 1.26$ |  | $-36.82 \pm 1.26$ | $-37.69 \pm 1.25$ | $57.49 \pm 1.37$ |
| RCFR-96% | $135.89 \pm 1.34$ | $495.46 \pm 1.85$ | $44.91 \pm 1.26$ | $149.31 \pm 1.41$ | $53.90 \pm 1.27$ | $10.17 \pm 1.25$ | $36.82 \pm 1.26$ |  | $0.00 \pm 1.25$ | $115.81 \pm 1.36$ |
| FULL | $139.90 \pm 1.33$ | $486.06 \pm 1.86$ | $47.98 \pm 1.27$ | $156.26 \pm 1.43$ | $56.77 \pm 1.27$ | $9.41 \pm 1.24$ | $37.69 \pm 1.25$ | $0.00 \pm 1.25$ |  | $116.76 \pm 1.36$ |

Table 1: One-on-one competition crosstable. Each cell is the bankroll for the row player in mb/h and 95% confidence interval of playing the row agent against the column agent for 30000000 hands where the starting player was changed after every hand. The table is partitioned according to abstraction or regressor complexity.

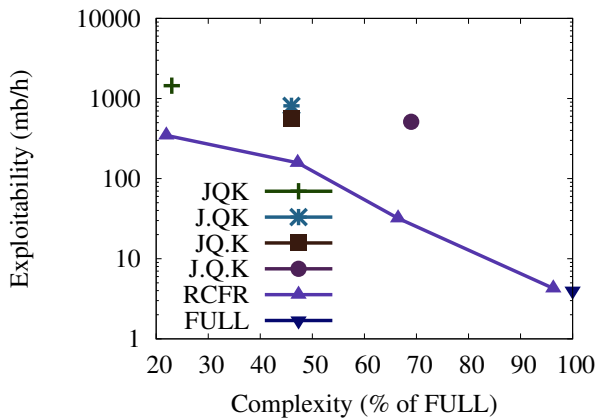

Figure 2: Exploitability of final strategies after 100000 iterations of RCFR, unabstracted CFR (`FULL`), or chance sampling CFR (`JQK`, `J.QK`, `JQ.K`, and `J.Q.K`). The horizontal axis shows the complexity of the solution method as a percentage of the size of the unabstracted game. RCFR's complexity is the size of the regression tree while the complexity of a conventional abstraction is the size of its abstract game.

team and a conventional agents team, the RCFR team wins 2033.34 mb/h in aggregate.

## Future Work

In this paper we introduced RCFR, a technique that obviates the need for abstraction as a preprocessing step by employing a regression algorithm online. The regressor essentially learns and tunes an abstraction automatically as the algorithm progresses greatly simplifying abstraction design.

The experimental results show this technique is quite promising. The next step is to scale to larger games, *e.g.*, no-limit Texas Hold'em, where abstraction is necessary. It is likely that RCFR is up to this challenge, but a number of interesting questions need be answered along the way.

First, CFR is the algorithm of choice for equilibrium-finding due to the powerful sampling schemes available to it. In this paper, we have not explored sampling with RCFR at all. Theoretically, there are no obvious restrictions that forbid sampling, but we posit that lower variance sampling schemes than are currently employed will be preferred.

Second, the choice of features and regressor are extremely important and remain open. Online regression algorithms are appealing due to their simplicity as well as removing the need to store data from past iterations. If this route is pursued, care must be taken to ensure access to strategy and updating the regressor remain inexpensive. This is possible, for example, by using sparse features.

Third, the average strategy is typically the result of the equilibrium computation. This average, too, must be encoded with a regressor in the case of large games. Conceptually, this poses no problems, but again care must be taken. In particular, due to the sequential nature of the policy, errors made by the regressor propagate and compound. The DAGGer algorithm corrects this by ensuring the regressor imposes the same distribution over future decisions as is observed in the true policy (Ross, Gordon, and Bagnell 2011).

## Acknowledgments

This work is supported by the ONR MURI grant N00014-09-1-1052, the National Sciences and Engineering Research Council of Canada (NSERC) and Alberta Innovative Technology Futures (AITF). Thanks to Compute Canada for computational resources and the Computer Poker Research Group (CPRG) for software infrastructure support.

# Appendix

Let $\{y^t\}_{t=1}^T$ be a sequence of reward vectors in $\mathcal{Y} \subset \mathbb{R}^N$.

Let $\phi(x) = \max\{0, x\}^2 = (x)_+^2$.

Let $\Phi(R) = \sum_{i=1}^N \phi(R_i)$.

Let $r^t = y^t - y^t \cdot x^t e$.

Let $R^T = \sum_{t=1}^T r^t$.

**Definition 2** (Regret-matching). *Let $x^{t+1} = (R^t)_+ / e \cdot (R^t)_+$.*

**Definition 3** (Blackwell's Condition).

$$\sup_{y \in \mathcal{Y}} \nabla \Phi(R^T) \cdot (y - y \cdot x^{t+1} e) \leq 0 \tag{1}$$

**Lemma 1.** *Regret-matching satisfies Blackwell's condition.*

*Proof.* For any $y$,

$$\nabla \Phi(R^T) \cdot (y - y \cdot x^{t+1} e) = 2(R^T)_+ \cdot (y - y \cdot ((R^t)_+ / e \cdot (R^t)_+) e) \tag{2}$$
$$= 2(y \cdot (R^T)_+ - y \cdot ((R^t)_+ / e \cdot (R^t)_+) e \cdot (R^T)_+) \tag{3}$$
$$= 2(y \cdot (R^T)_+ - y \cdot (R^T)_+) \tag{4}$$
$$= 0 \tag{5}$$

$\square$

**Theorem 6.** *If $\sup_{y \in \mathcal{Y}} \|y\|_2^2 \leq G$ then regret-matching has regret bounded by $\sqrt{GT}$.*

*Proof.* Adapted from Cesa-Bianchi and Lugosi.

by Lipschitz continuity of the gradient of $\Phi$

$$\Phi(R^{t+1}) \leq \Phi(R^t) + \nabla \Phi(R^t) \cdot (R^{t+1} - R^t) + \|R^{t+1} - R^t\|_2^2 \tag{6}$$
$$= \Phi(R^t) + \nabla \Phi(R^t) \cdot r^{t+1} + \|r^{t+1}\|_2^2 \tag{7}$$

by Blackwell's condition

$$\leq \Phi(R^t) + \|r^{t+1}\|_2^2 \tag{8}$$
$$\leq \Phi(R^t) + G \tag{9}$$

iterating the inequality

$$\leq (t+1)G \tag{10}$$
$$\max_{i \in [N]} R_i^T \leq \max_{i \in [N]} (R_i^T)_+ = \|(R^T)_+\|_\infty \tag{11}$$
$$\leq \|(R^T)_+\|_2 = \sqrt{\Phi(R^T)} \tag{12}$$
$$\leq \sqrt{TG} \tag{13}$$

$\square$

Suppose Blackwell's condition does not hold exactly. In particular, if

$$\sup_{y \in \mathcal{Y}} \nabla \Phi(R^t) \cdot (y - y \cdot x^{t+1} e) \leq \epsilon^{t+1} \tag{14}$$

Then using the above proof structure, we instead get a regret bound

$$\max_{i \in [N]} R_i^T \leq \sqrt{TG + \sum_{t=1}^T \epsilon^t} \tag{15}$$

That is, so long as our choice of $x^t$ ensures $\sum_{t=1}^{T} \epsilon^t$ grows less than $T^2$ our algorithm will be no-regret. This result is in Jafari et. al..

Consider letting $x^{t+1} \propto (\tilde{R}^T)_+$.

$$\nabla \Phi(R^T) \cdot (y - y \cdot x^{t+1} e) = 2(R^T)_+ \cdot (y - y \cdot ((\tilde{R}^T)_+/e \cdot (\tilde{R}^T)_+)e) \tag{16}$$
$$= 2y \cdot ((R^T)_+ - ((\tilde{R}^T)_+/e \cdot (\tilde{R}^T)_+)e \cdot (R^T)_+) \tag{17}$$

by Cauchy-Schwarz

$$\leq 2\|y\|_2 \|(R^T)_+ - ((\tilde{R}^T)_+/e \cdot (\tilde{R}^T)_+)e \cdot (R^T)_+\|_2 \tag{18}$$

Note that the rightmost norm is almost $\|(R^T)_+ - (\tilde{R}^T)_+\|_2$, except we have scaled $(\tilde{R}^T)_+$ to have the same $L1$-norm as $(R^T)_+$. Projections preserve distances, therefore

$$\nabla \Phi(R^T) \cdot (y - y \cdot x^{t+1} e) \leq 2\|y\|_2 \|(R^T)_+ - (\tilde{R}^T)_+\|_2 \tag{19}$$

If $\|R^T - \tilde{R}^T\|_2 \leq \epsilon$, then regression regret-matching has regret bounded by $\sqrt{TG} + 2T\sqrt{G}\epsilon$.

If $\|R^T - \tilde{R}^T\|_2/T \leq \epsilon$, then regression regret-matching has regret bounded by $\sqrt{TG} + 2T^2\sqrt{G}\epsilon$.

Note that $\|x\|_2 \leq \sqrt{N}\|x\|_\infty$. That is, $\sqrt{G} \leq \sqrt{NL}$.